\documentclass[letterpaper, 10 pt, conference]{ieeeconf} 
\IEEEoverridecommandlockouts                             
\usepackage{color}
\usepackage{graphicx} 
\usepackage{mathtools} 
\usepackage{siunitx}
\usepackage{dblfloatfix}

\usepackage{adjustbox}
\usepackage{array}

\newcolumntype{R}[2]{%
    >{\adjustbox{angle=#1,lap=\width-(#2)}\bgroup}%
    l%
    <{\egroup}%
}
\newcommand*\rot{\multicolumn{1}{R{45}{1em}}}

\title{\LARGE \bf
Soft Everting Prosthetic Hand and Comparison with \\ Existing Body-Powered Terminal Devices
}

\author{Gayoung Park$^{1*}$, Katalin Schäffer$^{1,2*}$, and Margaret M. Coad$^{1}$
\thanks{$^{1}$Department of Aerospace and Mechanical Engineering, University of Notre Dame, Notre Dame, IN 46556, USA. {\tt\small \{gpark2, kschaff2, mcoad\}@nd.edu}}%
\thanks{$^{2}$Faculty of Information Technology and Bionics, Pázmány Péter Catholic University, 1083 Budapest, Hungary.}%
\thanks{$^{*}$Indicates equal contribution.}%
} 

\begin{document}

\maketitle
\thispagestyle{empty}
\pagestyle{empty}

\begin{abstract}
In this paper, we explore the use of a soft gripper, specifically a soft inverting-everting toroidal hydrostat, as a prosthetic hand. We present a design of the gripper integrated into a body-powered elbow-driven system and evaluate its performance compared to similar body-powered terminal devices: the Kwawu 3D-printed hand and the Hosmer hook. Our experiments highlight advantages of the Everting hand, such as low required cable tension for operation (1.6 N for Everting, 30.0 N for Kwawu, 28.1 N for Hosmer), limited restriction on the elbow angle range, and secure grasping capability (peak pulling force required to remove an object: 15.8 N for Everting, 6.9 N for Kwawu, 4.0 N for Hosmer).  
In our pilot user study, six able-bodied participants performed standardized hand dexterity tests. 
With the Everting hand compared to the Kwawu hand, users transferred more blocks in one minute and completed three tasks (moving small common objects, simulated feeding with a spoon, and moving large empty cans) faster (p~$\leq$~0.05). With the Everting hand compared to the Hosmer hook, users moved large empty cans faster (p~$\leq$~0.05) and achieved similar performance on all  other tasks.
Overall, user preference leaned toward the Everting hand for its adaptable grip and ease of use, although its abilities could be improved in tasks requiring high precision such as writing with a pen, and in handling heavier objects such as large heavy cans.
\end{abstract}

\section{Introduction} \label{sec:Introduction}
For individuals with limb difference that affects their hand function, prosthetic hands have the potential to restore their ability to achieve everyday tasks \cite{cordella2016literature, stephens2019survey}. A large variety of hand designs have been developed with promising functional capabilities. However, when we examine how these solutions are integrated into user-ready devices, there remains significant demand for designs that are low-cost, light-weight, quick to repair, and easy to use.  Body-powered solutions, such as the Hosmer hook or grip prehensors, continue to be popular despite their non-anthropomorphic aesthetics and simpler functionality compared to humanoid hands ~\cite{resnik2021longitudinal,kim2022influence}. This reflects user preferences for simplicity and practicality.

Balancing functionality with simplicity has motivated researchers to reduce the complexity of prosthetic hand designs while meeting user needs. Many solutions implement an underactuated design~\cite{vertongen2020mechanical}, which makes precise grip positions hard to achieve. Adding compliance to these mechanisms has helped mitigate this challenge by increasing adaptability~\cite{carrozza2004spring, dollar2006robust}. 
One example of a low-cost, customizable approach is the 3D-printed e-NABLE Kwawu hand, which achieves a voluntary-closing grip through body-powered actuation~\cite{kwawu_hand}. As 3D printing continues to advance, its flexibility and low fabrication costs further support a trend toward customizable, accessible prosthetic hand designs~\cite{kim2022influence,vertongen2020mechanical}. 


Innovations in soft robotics have introduced new possibilities for prosthetic hands that prioritize adaptability through large body deformations~\cite{shintake2018soft}. For example, universal grippers use either granular jamming~\cite{brown2010universal} or magnetorheological fluid~\cite{nishida2016development} to conform to objects by flowing around them and locking into place, and grippers made of soft fluidic elastomer bending actuators conform to an object when pressurized~\cite{ilievski2011soft, hao2016universal}. Another promising concept is inversion and eversion of a soft toroidal hydrostat, i.e., turning outside-in and inside-out a flexible and slightly stretchable membrane that is formed into a toroidal shape and filled with liquid. Researchers have demonstrated the ability of this type of gripper to pick up objects including eggs, screwdrivers, cloth, hair~\cite{li2020bioinspired}, and even a sheet of paper lying flat on the table~\cite{sui2022bioinspired}, and to gently convey objects inside~\cite{root2021bio}. A key advantage of this design is its closed volume structure, which maintains internal fluid pressure without external pumps. 

\begin{figure}[t]
    \centering
    \includegraphics[width=\columnwidth]{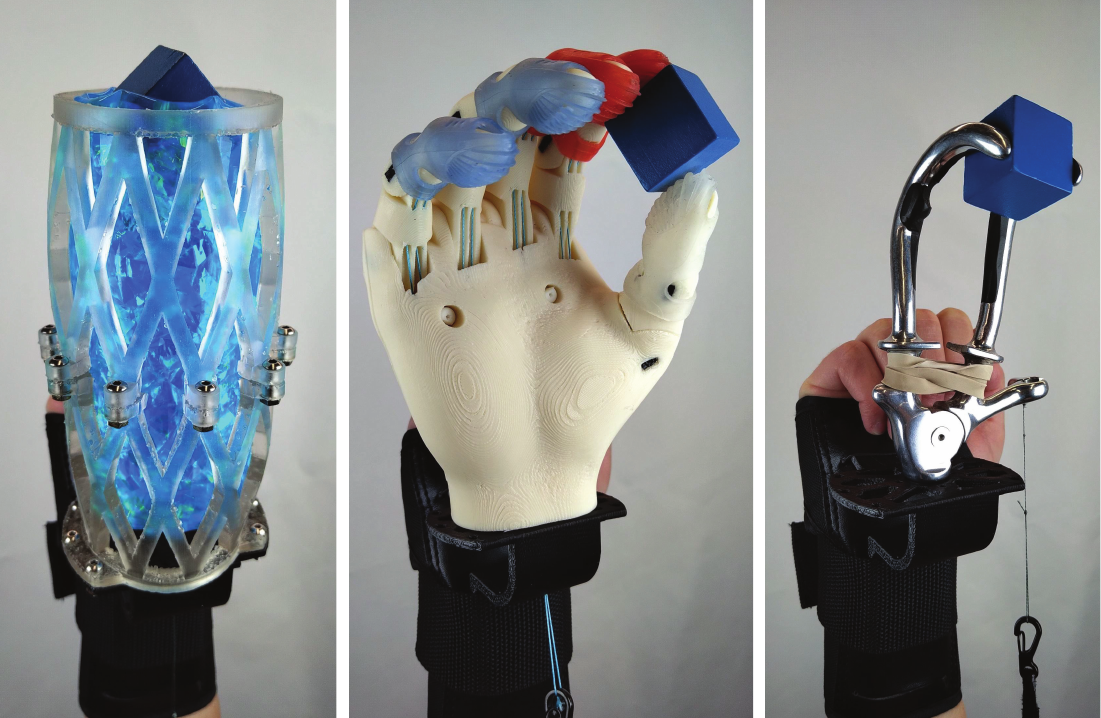}
    \caption{Body-powered prosthetic hand designs with different terminal devices: our Everting hand design (left), the 3D printed Kwawu hand~\cite{kwawu_hand} from \mbox{e-NABLE}~\cite{enable} (middle), and the Hosmer hook (right).}
    \label{fig:GlamorShot}
    \vspace{-0.5 cm}
\end{figure}

\begin{table*}[b]
\caption{The Three Examined Terminal Devices with Body-Powered Transmission\label{tab:three_designs}}
\vspace{-0.2 cm}
\label{specs}
\centering
\begin{tabular}{|>{\centering\arraybackslash}p{1.6cm}|>{\centering\arraybackslash}p{1.5cm}|>{\centering\arraybackslash}p{4.0cm}|>{\centering\arraybackslash}p{2.8cm}|>{\centering\arraybackslash}p{2.3cm}|>{\centering\arraybackslash}p{2.5cm}|}
\hline
\textbf{Name} & \textbf{DoFs}& \textbf{Structure and compliance} & \textbf{Opening/closing style} & \textbf{Weight}  & \textbf{Aesthetics}\\
\hline
\hline
Everting hand (our design) & infinite \newline (1 actuated) & soft inverting-everting toroidal hydrostat forms around an object & voluntary closing & 355.0 g \newline (279.4 g hydrostat) & non-anthropomorphic\\
\hline
Kwawu hand & 9 \newline (1 actuated) & 3D printed links and flexible joints, cable-driven underactuated & voluntary closing & 224.4 g & anthropomorphic\\
\hline
Hosmer hook & 1 \newline (1 actuated) & two rigid links with an elastic band closing the gripper & voluntary opening & 99.3 g  &  non-anthropomorphic\\
\hline
\end{tabular}\\
\vspace{-0.0 cm}
\end{table*}

In this study, we explore the use of a soft inverting-everting toroidal hydrostat as a body-powered prosthetic hand and compare it with other body-powered prosthetic devices: the 3D-printed Kwawu hand and the Hosmer hook (Fig.~\ref{fig:GlamorShot}). The everting hydrostat’s flexibility and secure, enclosing grasp make it well-suited for various object shapes and sizes. We developed a body-powered adapter that allows able-bodied users to operate the Everting hand and other terminal devices using a simple cable-driven transmission system actuated by elbow movement. We evaluate all devices by measuring, first, the cable tension required to operate each prosthetic terminal device, and second, the pulling force necessary to remove an object from their grasp. Additionally, we conduct a pilot user study with six able-bodied participants performing hand function tests~\cite{mathiowetz1985adult, jebsen1969objective} to gain more insights about the advantages and limitations of the three prosthetic terminal device designs for performing everyday tasks.


\section{Design} \label{sec:Design}
In this section, we present the design considerations and implementation details for our soft everting prosthetic device, along with the description of the two other terminal devices used for comparison.

\begin{figure}[tb]
    \centering
    \includegraphics[width=\columnwidth]{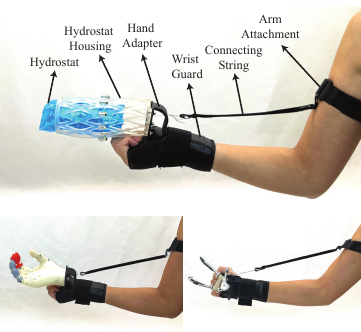}
    \caption{The three prosthetic terminal devices with body-powered actuation. Our Everting hand is attached to the right hand of an able-bodied user through a custom-designed prosthetic simulator, and it is connected to the body powered transmission system (top). The Kwawu hand (bottom left) and the Hosmer hook (bottom right) connect to the same prosthetic simulator and transmission system.}
    \label{fig:Design}
    \vspace{-0.5cm}
\end{figure}

\subsection{Everting hand}

The top image in Fig.~\ref{fig:Design} shows the implementation of our Everting hand design in our body-powered system. The hand comprises the hydrostat, a flexible housing to prevent collapse, and a stopper inserted inside the hydrostat to provide a fixed point for inversion and eversion control. For the soft toroidal hydrostat, we used a commercially available toy called a ``water wiggler" (Water Snake Wiggler, Super Z Outlet, California, USA). Its length is 12.7~cm, its outer diameter is 5.1~cm, and its weight is 279.4~g. The housing for the hydrostat is a flexible mesh with diagonal crossing pattern forming a cylinder (14.5~cm long and 5.2~cm inner diameter) to support the hydrostat to keep it from collapsing. The structure was assembled from two segments which were 3D printed using flexible resin (Flexible 80A resin, Formlabs) and attached by bolts in the middle line of the cylinder. The length of the housing is slightly longer than the length of the hydrostat so that it can provide the hydrostat enough space to travel as it inverts and everts. To make it possible to invert the hydrostat by pulling on a thread, we placed a stopper inside the hydrostat, which is held in place by frictional force. The stopper is a 3D printed dumbbell shaped part with a diameter of 1.4~cm and a height of 1.8~cm. The weight of the complete device is 355.0~g.

\subsection{Body-Powered Transmission System}
We designed an adapter that connects the Everting hand and other terminal devices to the forearm of an able-bodied user, shown in Fig.~\ref{fig:Design}. This adapter includes an off-the-shelf wrist guard (Wrist Guards, Tanden LLC, China) with a modified splint that we 3D-printed in ABS-M30 material (Stratasys). The splint has an extended area perpendicular to the forearm where the base of each terminal device can be securely attached via a screw and a nut.

The body-powered actuation system consists of a connecting cable (PowerPro Microfilament Braided Line, 40 lb, Shimano North America Fishing, Inc., South Carolina, USA) and a hook and loop band for the upper arm. The cable connects the terminal device’s actuator cable to the arm attachment, translating elbow joint movement into device actuation.

\subsection{Other Prosthetic Terminal Devices}
To compare performance, we used two additional terminal devices: a 3D-printed right-handed Kwawu hand and a Hosmer hook (Fig.~\ref{fig:Design}). A summary of all three devices and their structural characteristics is presented in Table~\ref{tab:three_designs}.

The Kwawu hand, an e-NABLE design~\cite{kwawu_hand}, was selected for its compatibility with elbow-powered actuation. Our version of the Kwawu hand was scaled in OpenSCAD~\cite{openscad} based on a hand breadth size of 8.3~cm. This hand size corresponds to the mean of the average woman and average man hand breadth reported in~\cite{nasa_anthropometric_source_book}, and thus, represents a generic size that fits multiple users. The hand is composed of two rigid 3D printed links (ABS-M30 material, Stratasys) for each of the five fingers,  0.2~cm thick flexible hinges (Economical Abrasion-Resistant SBR Rubber, McMaster-Carr Part 8634K332, Indiana, USA) for the joints, a hollow base composed of a top and a palmar 3D printed part, a pulley system inside the hand, and a screw (1/2"-20 Thread Size, 3/4" Long Stainless Steel Hex Head Screw, McMaster-Carr Part 93190A378, Indiana, USA) at the base of the hand. Fishing lines were routed through the fingers and connected through a pulley system, resulting in a voluntary closing, under-actuated mechanism. Soft fingertip covers (Lee Tippi Micro Gel Fingertip Grips, Lee Products Company, Minnesota, USA) were placed on the hand for improved grip. The total weight of the assembled Kwawu hand is 224.4~g.

The third terminal device we used is a 12~cm long right-handed aluminum Hosmer Dorrance prosthetic hook (Fillauer, Model 88X), which weighs 99.3~g. It is a canted split hook with a single hinge joint that can be voluntarily opened through pulling on an attached cable while opposing an elastic band. For the purpose of this study, we replaced the original band with a weaker rubber band (Advantage Rubber Bands Size 64, Alliance, Arcansas, USA), such that the hook can be operated with our custom body-powered transmission system.

\section{Experiments} \label{sec:Experiment}
 We conducted two experiments to measure and compare the performance of the three terminal devices. The first experiment measured the cable tension required to operate each device, providing an indication of the demand placed on users. The second experiment measured the pulling force necessary to remove an object from each device's grasp, assessing its grasp security.

\subsection{Cable Tension Measurement}
\subsubsection{Setup}
In this experiment, our goal was to measure the cable tension generated at the arm attachment point of the body-powered transmission system as each terminal device was actuated through elbow movement. 

To standardize measurements, we used a mannequin arm with a length of 71.1~cm and securely attached a force sensor (Nano17, ATI) to the upper arm in series with the connecting cable. The terminal device was attached to the arm using the prosthetic hand adapter. A goniometer was attached to the arm to monitor the elbow angle, allowing for precise tracking of the joint’s extension. 

Each terminal device was connected to the adapter on the mannequin arm, and the arm was positioned so that the forearm aligned horizontally on a tabletop. Starting from a 90° elbow angle, we manually extended the elbow to the maximum angle, recording cable tension at each 10° increment. This process was repeated three times per device to ensure measurement reliability. For the Everting hand, additional measurements were taken with the forearm in a vertical configuration and the hand pointing downward to assess the effect of orientation on cable tension. The experimental setup in the horizontal configuration is shown in Fig.~\ref{fig:CableTension} with the Everting hand.

\subsubsection{Results}
\begin{figure}[tb]
    \centering
    \includegraphics[width=\columnwidth]{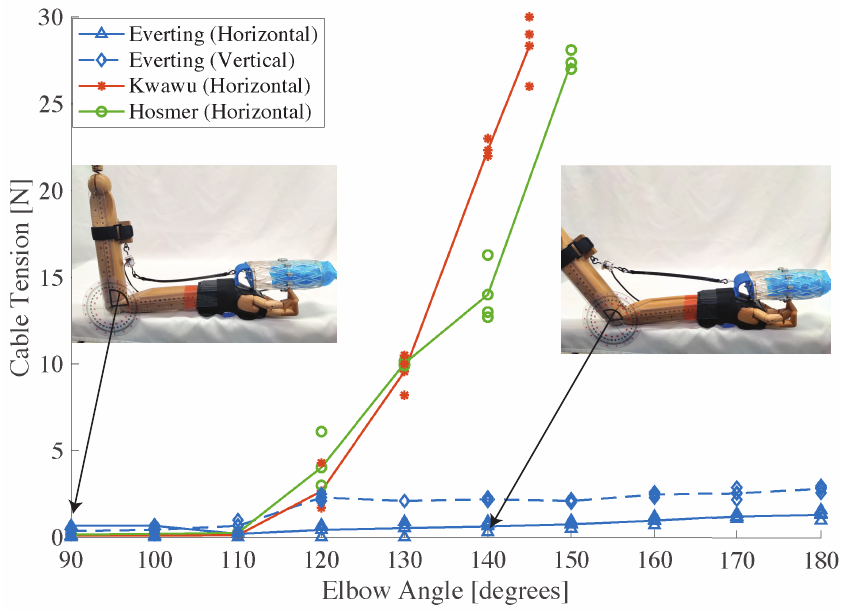}
    \caption{Measurement setup and experimental results of the cable tension measurement. The plot shows the measured cable tensions as a function of elbow angle for the three terminal devices and, in the case of the Everting hand, also for two different orientations. The photos illustrate the measurement setup with the mannequin arm in the horizontal configuration and with the Everting hand as the terminal device. A force sensor is connected to the actuator cable at the arm attachment to record the cable tension at different elbow angles. The Everting hand requires lower cable tension than the other two devices, and it can operate over a larger range of elbow angles.}
    \label{fig:CableTension}
    \vspace{-0.5cm}
\end{figure}

 As shown in Fig.~\ref{fig:CableTension}, tension increased with elbow extension, particularly beyond 110$^\circ$. When the cable is stretched, the Everting hand requires a static, low tension force to invert (peak tension of 1.6~N), while the other devices require gradually increased tension when the elbow extends and reach peak tensions of 30.0~N for the Kwawu hand and 28.1~N for the Hosmer hook.
 For the Everting hand, although the unsupported weight of the hydrostat adds to the cable tension in the vertical orientation, the peak tension does not exceed 3.0~N. As a consequence of the observed force profile, the Everting hand does not limit the range of elbow extension, while the other devices stop the elbow extension at around 150$^\circ$. In these tests, the tip of the everting hydrostat was 2.5~cm outside the housing when cable tension was not applied and the stopper was close to the tip of the hydrostat. When the elbow reached the maximum angle of 180$^\circ$, the hydrostat was close to but had not fully reached the base of the housing.

\subsection{Grasping Measurement}

\begin{figure}[tb]
    \centering
    \includegraphics[width=\columnwidth]{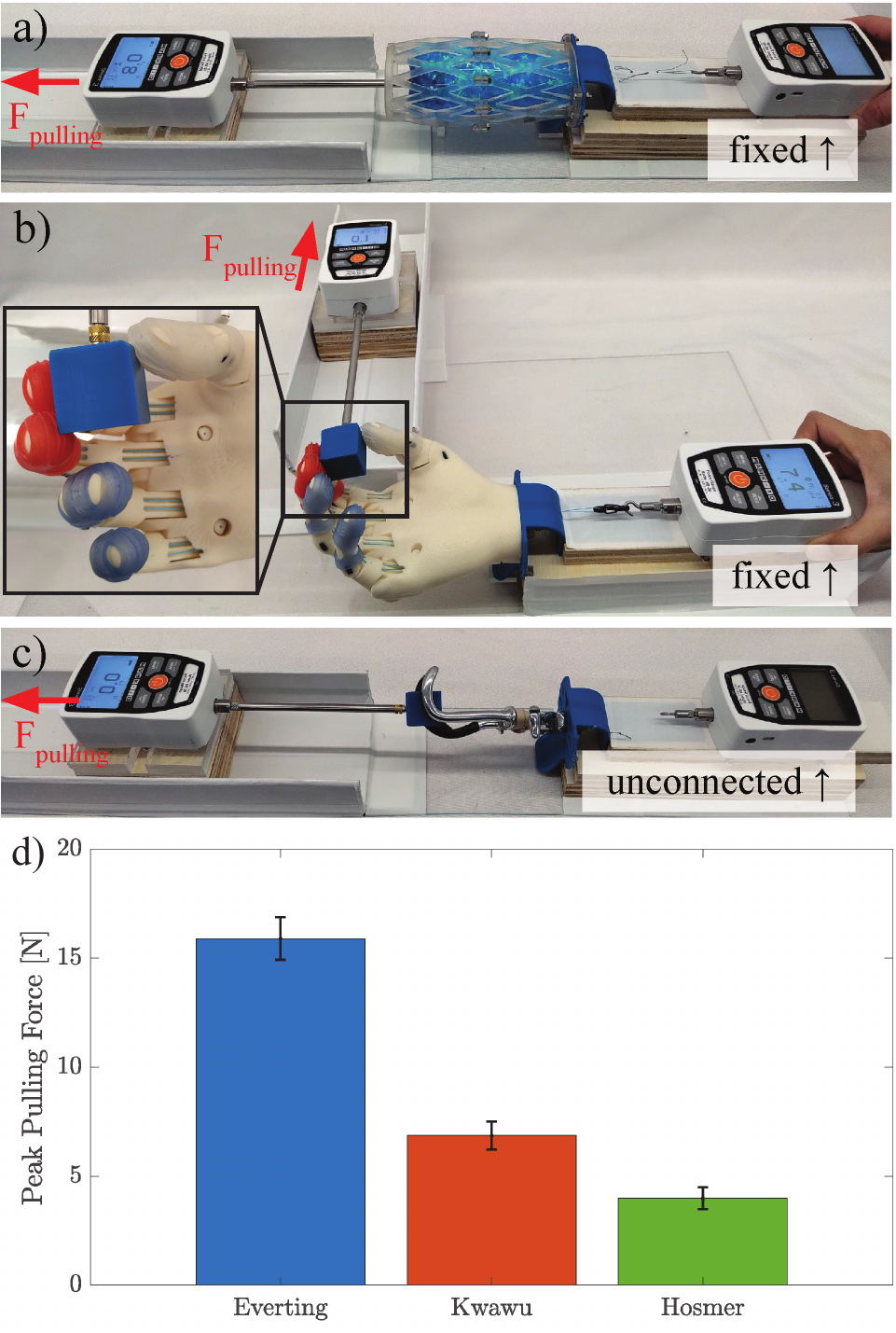}
    \caption{Experimental setup and results of grasp force measurement. (a-c) The measurement setup for the Everting hand, Kwawu hand, and Hosmer hook, respectively. The setup was used to grasp an object (blue cube) with each terminal device and measure the pulling force (indicated with red arrows) required to pull the object out of the grasp. (d) The mean and the standard deviation of six measurements for each of the devices, showing that the Everting hand resisted higher peak force than the other two.}
    \label{fig:GraspingMeasurement}
    \vspace{-0.5cm}
\end{figure}

\subsubsection{Setup}
This experiment was designed to evaluate the grasping performance of each terminal device by measuring the pulling force necessary to remove a small object from the device’s grasp. We attached a 3D-printed cube (Blue Tough PLA material, Ultimaker) with 2.5~cm edges to a force sensor (M3 Series, MARK-10) and measured the peak pulling force when the object comes out of the grasp. Each terminal device was fixed in place by attaching the prosthetic hand adapter to a stand to ensure consistent positioning. 

For the Everting hand (Fig.~\ref{fig:GraspingMeasurement}(a)) and Hosmer hook (Fig.~\ref{fig:GraspingMeasurement}(c)), we applied force from the front of the device, parallel to the hypothetical arm direction, as this was their easiest way to come out of the grasp. For the Kwawu hand (Fig.~\ref{fig:GraspingMeasurement}(b)), we applied force from the side, as its primary pinch grasp was weakest in this orientation.

For the Everting hand, the average value of the pulling force that we applied to the cable with the MARK-10 when the hydrostat fully enclosed the cube was 2.8~N. This cable tension increased and reached an average of 15.2~N of peak tension when the cube came out of the grasp. For the Kwawu hand, we applied the highest force (20$\pm$0.3~N) that we could manually hold steadily during the measurement. Since the Hosmer hook is a voluntary opening device, it did not require applied cable tension, and the grasping force depended only on the elasticity of the rubber band.

\subsubsection{Results}
Fig.~\ref{fig:GraspingMeasurement}(d) shows the average pulling force each device resisted. The Everting hand demonstrated the highest resistance, requiring an average pulling force of 15.8~N, which indicates strong grasp security. The enclosing grasp design and high-friction hydrostat surface contributed to this secure hold. However, the Kwawu hand showed significantly smaller resistance to the pulling force, reaching only an average of 6.9~N peak force with the applied pinch grasp (Fig.~\ref{fig:GraspingMeasurement}(b)). Regarding the user effort, the Kwawu hand requires a  constantly applied, high cable tension to maintain a strong grasp. 
In contrast with the other terminal devices, the Hosmer hook does not require any cable tension to maintain the grasp, but its resistance to the pulling force was the smallest, with an average of 4.0~N recorded peak force. The grasp could be improved with a stronger elastic band or improved surface properties, as the hook is the only examined terminal device that does not have a covering at the grasping surface. 

\section{User Study} \label{sec:UserStudy}

 To assess the usability, dexterity, and user preferences for each prosthetic terminal device, we conducted a user study with six able-bodied participants. Each participant used the Everting hand, Kwawu hand, and Hosmer hook to perform manipulation tasks, providing a comparative perspective on each device’s functionality in real-life scenarios.

\subsection{User Study Procedure}
\begin{figure}[tb]
    \centering
    \includegraphics[width=\columnwidth]{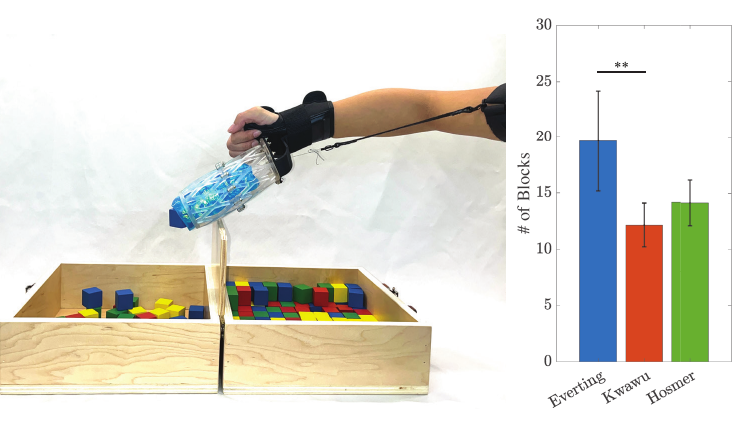}
    \caption{Setup for the Box and Blocks Test with the Everting hand (left), and results of the user study participants using the three terminal devices (right). Error bars indicate the standard deviation among users, and statistical significance at the level of p~$\leq$~0.01 in a post-hoc test is denoted with two asterisks. The Everting hand had the highest block transfer rate, highlighting the advantage of its adaptable grip in repetitive movements, while the Kwawu hand had the lowest rate due to alignment and grip control challenges. }
    \label{fig:BoxAndBlocksTest}
    \vspace{-0.5cm}
\end{figure}

\begin{figure*}[tb]
    \centering
    \includegraphics[width=\textwidth]{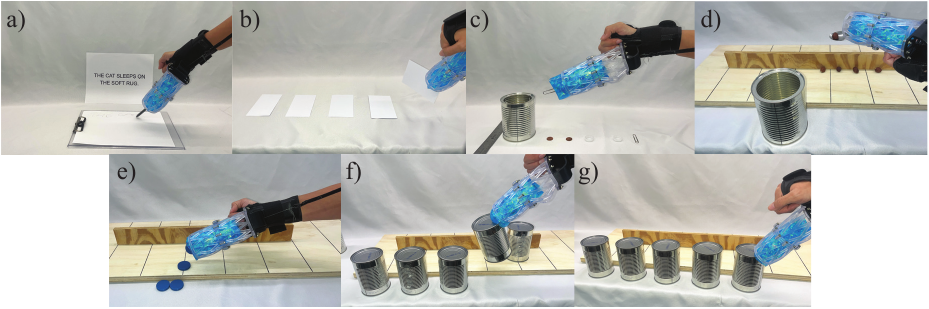}
    \caption{Photos of a user completing the seven tasks of the Jebsen-Taylor Test with the Everting hand: (a) writing a sentence with a pen, (b) turning over cards, (c)  picking up small objects (coins, bottle caps, and paper clips), (d) simulated feeding with a spoon, (e) stacking checkers, (f) moving large, light objects, and (g) moving large, heavy objects.}
    \label{fig:JebsenTasks}
    \vspace{-0.5cm}
\end{figure*}

We selected two standardized hand dexterity tests commonly adapted from clinical assessments. The first was the Sammons Preston Box and Block Test (Box and Blocks Test)~\cite{mathiowetz1985adult}. The test (illustrated in Fig.~\ref{fig:BoxAndBlocksTest} with the Everting hand) evaluates manual dexterity by measuring the ability to move 2.5~cm cubic wooden blocks from one compartment to another within a one-minute time interval. The second was the Jebsen-Taylor Hand Function Test (Jebsen-Taylor Test)~\cite{jebsen1969objective} (Fig.~\ref{fig:JebsenTasks}), which is another standardized hand function assessment that involves a series of seven timed tasks with diverse objects and manipulation requirements, including writing a sentence, turning over cards, picking up small items (e.g., coins, bottle caps, paper clips), simulated feeding with a spoon, stacking checkers, and moving light and heavy cans. We classified the tasks as failed if not completed within two minutes, and we assigned a completion time of 120 seconds for failed tasks.

The user study was approved by the University of Notre Dame's Institutional Review Board (IRB) under protocol number 24-10-8867. Six participants, aged 21 to 24 (two female, four male), all right-handed with normal upper limb function, were recruited from the university's student body. After being fitted with each device, participants were given a few minutes to become familiar with its operation. They completed the Blocks and Box Test first and the Jebsen-Taylor Test second, while their performance was recorded. Afterward, they removed the prosthetic hand and filled out a survey form about their experience using it. Each participant then repeated the same procedure with the other two terminal devices. Finally, the participant completed a survey comparing their experiences with all three devices. The order in which the three devices were presented to each participant was randomized to prevent learning bias. 

\subsection{Statistical Analysis}

We conducted statistical analyses to evaluate performance differences across the three devices in both the Box and Blocks Test and the Jebsen-Taylor Test. For both assessments, we used the Kruskal-Wallis statistical test due to its suitability for non-normally distributed data, followed by post-hoc analysis using the Dunn-Sidak method for pairwise comparisons when significant differences were identified. 

\subsection{Box and Blocks Test Results}
Fig.~\ref{fig:BoxAndBlocksTest} shows the Box and Blocks Test results of our user study. The Everting hand produced the highest block transfer rate, with participants averaging 19.7 blocks per minute, indicating that its adaptable grip design supports repetitive and stable movements.

The Kwawu hand had the lowest average transfer rate, with participants averaging 12.2 blocks per minute. Participants reported difficulty with alignment and grip control for each transfer. Survey feedback suggested that the Kwawu hand’s limited grip adaptability posed challenges for the rapid, repetitive movements required in the test.

The Hosmer hook performed moderately, with participants averaging 14.2 blocks per minute. While this hand was effective for repetitive tasks, some participants needed to adjust grip positioning occasionally, which may have reduced efficiency.

The Kruskal-Wallis test yielded a statistically significant difference among the devices (p~=~0.007). The post-hoc analysis confirmed that the Everting hand significantly outperformed the Kwawu hand (p~=~0.005), but no significant differences were found between the Everting hand and the Hosmer hook (p~=~0.15) or between the Kwawu hand and the Hosmer hook (p~=~0.54). Thus, while the Everting hand provided a clear advantage over the Kwawu hand, its performance was comparable to that of the Hosmer hook in terms of manual dexterity.

\subsection{Jebsen-Taylor Test Results}
\begin{figure}[tb]
    \centering
    \includegraphics[width=\columnwidth]{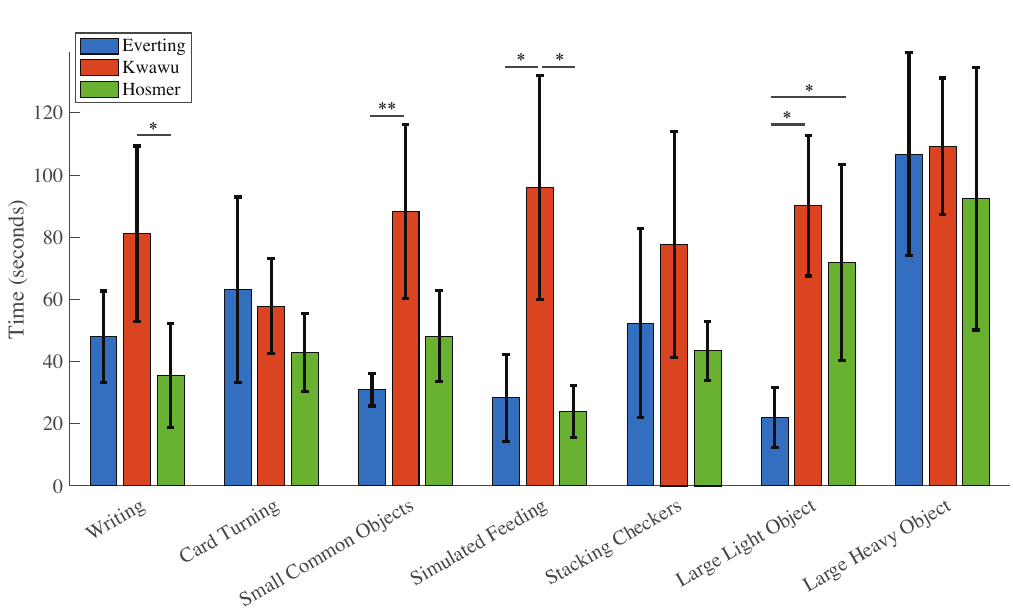}
    \caption{Jebsen-Taylor Test results for the six user study participants. Bars denote average completion time across all users for all seven tasks with the three terminal devices. Error bars denote the standard deviation among users, and asterisks denote statistically significant differences (one asterisk for post-hoc test values of p~$\leq$~0.05 and two asterisks for p~$\leq$~0.01). The Everting hand performed similarly to the Kwawu hand on some tasks, but significantly outperformed it on \textit{Small Common Objects}, \textit{Simulated Feeding}, and \textit{Large Light Objects}. The Everting hand performed similarly to the Hosmer hand on most tasks, except \textit{Large Light Objects}, where the Everting hand was significantly better.  
    }
    \label{fig:JebsenResults}
    \vspace{-0.5cm}
\end{figure}

Fig.~\ref{fig:JebsenResults} presents the Jebsen-Taylor Test results of our user study, and 
Table~\ref{tab:Jebsen_fails} displays the number of participants who could not complete each task with each device.

The Everting hand was particularly effective in tasks involving irregularly shaped or large objects, such as moving \textit{Small Common Objects} and \textit{Large Light Objects}. However, in the \textit{Large Heavy Objects} task, only one out of six participants successfully completed the task using the Everting hand. In tasks requiring precise control, such as \textit{Writing} and \textit{Stacking Checkers}, the Everting hand faced limitations. Participants noted that the fluid-filled structure created a slight shakiness, which impacted the device's stability in fine motor tasks such as \textit{Writing}. Also, the structure restricted visibility within the grip area, making it difficult to see the precise positioning of smaller items like checkers.

The Kwawu hand showed the longest average completion times across most tasks. Grip precision and stability limitations were especially apparent in tasks like \textit{Simulated Feeding}, \textit{Stacking Checkers}, and \textit{Writing}, with participants often taking more time or failing to complete tasks within the time limit. Feedback indicated that the Kwawu hand’s underactuated mechanism and alignment challenges reduced overall task efficiency.

The Hosmer hook demonstrated consistent and moderate performance across most tasks. It showed particular strength in tasks requiring fine motor control, such as \textit{Card Turning} and \textit{Stacking Checkers}. Its stable, straightforward design allowed participants to securely handle \textit{Small Common Objects}. Limitations in grip strength and stability affected performance in tasks requiring manipulation of \textit{Large Light Objects} and \textit{Large Heavy Objects}, resulting in longer completion times.

\begin{table}[b]
\caption{Number of failed task executions among six participants\label{tab:Jebsen_fails}}
\vspace{-0.2 cm}
\label{failures}
\centering
\begin{tabular}{|>{\centering\arraybackslash}p{0.9cm}|>
{\centering\arraybackslash}p{0.5cm}|>
{\centering\arraybackslash}p{0.5cm}|>
{\centering\arraybackslash}p{0.5cm}|>
{\centering\arraybackslash}p{0.5cm}|>
{\centering\arraybackslash}p{0.5cm}|>
{\centering\arraybackslash}p{0.5cm}|>
{\centering\arraybackslash}p{0.5cm}|>
{\centering\arraybackslash}p{0.1cm}|}
\hline
\textbf{Device} & \rot{\textbf{Writing}} & \rot{\textbf{Cards}} & \rot{\textbf{Small Obj.}} & \rot{\textbf{Feeding}}  & \rot{\textbf{Checkers}} & \rot{\textbf{Light Obj.}} & \rot{\textbf{Heavy Obj.}} & \\
\hline
\hline
 Everting &0& 1& 0& 0& 0&0 &5& \\
\hline
Kwawu &1& 0& 2& 3& 1 &1 &3 &\\
\hline
Hosmer &0& 0& 0& 0& 0&0 &3 &\\
\hline
\hline
\end{tabular}\\
\vspace{-0.0 cm}
\end{table}

The Kruskal-Wallis test identified significant differences between the three devices in several tasks: \textit{Writing} (p~=~0.02), \textit{Small Common Objects} (p~=~0.005), and \textit{Large Light Objects} (p~=~0.007), with an almost significant difference in \textit{Simulated Feeding} (p~=~0.06). Post-hoc analysis showed that in the \textit{Writing} task, the Hosmer hook performed significantly better than the Kwawu hand (p~=~0.01), with no significant difference between the Everting hand and the Hosmer hook (p~=~0.47). For the \textit{Small Common Objects} task, the Everting hand significantly outperformed the Kwawu hand (p~=~0.005), with no significant difference between the Everting hand and the Hosmer hook (p~=~0.13). In \textit{Simulated Feeding}, the Everting hand showed a significant advantage over the Kwawu hand (p~=~0.03), and the Kwawu hand was significantly slower than the Hosmer hook (p~=~0.01), while no significant difference was found between the Everting hand and the Hosmer hook (p~=~0.98). Finally, in the \textit{Large Light Objects} task, the Everting hand outperformed both the Kwawu hand (p~=~0.01) and the Hosmer hook (p~=~0.04). Tasks \textit{Card Turning}, \textit{Stacking Checkers}, and \textit{Large Heavy Objects} did not show statistically significant differences across the devices (p-values of 0.22, 0.28, and 0.38, respectively). Together, these results highlight the Everting hand’s strengths in specific tasks, particularly those involving smaller or lighter objects.

\subsection{Participant Preferences}
Each participant ranked the three prosthetic devices based on their overall experience. Three chose the Everting hand as their first preference, while three ranked the Hosmer hook first. All participants ranked the Kwawu hand as their least preferred device. The Everting hand was favored for its adaptability in grasping various object shapes and sizes and its capabilities in tasks involving larger objects. However, limited visibility within the grip area and shakiness impacted performance in the \textit{Stacking Checkers} and \textit{Writing} tasks. With the Kwawu hand, participants cited difficulties in grip control and alignment, particularly in tasks requiring precise grasping, such as \textit{Writing} and handling \textit{Small Common Objects}. The Hosmer hook was rated as intuitive and effective for tasks requiring fine manipulation, such as \textit{Writing} and \textit{Card Turning}, although grip strength limitations were noted for larger items.

\section{Conclusions} \label{sec:Conclusion}
We explored the use of a soft inverting-everting toroidal hydrostat that we implemented as a prosthetic hand within a body-powered system. We designed a device that utilizes the adaptable grip of the hydrostat and evaluated its performance along with two other body-powered devices, the Kwawu 3D-printed hand and the Hosmer hook.

Our experiments demonstrated the unique benefits of the Everting hand in that it required a notably low peak cable tension of 1.6~N for operation, compared to 30.0~N and 28.1~N for the Kwawu and Hosmer devices. This low tension requirement suggests a reduced demand on users, and this allows for a broader angle range of elbow extension and more comfortable operation. Additionally, the Everting hand demonstrated secure grasping, where it resisted a pulling force of 15.8~N—significantly higher than the 6.9~N and 4.0~N resisted by the Kwawu hand and Hosmer hook, respectively.

The pilot user study confirmed the Everting hand’s effectiveness in practical applications, as it consistently performed well in tasks involving varied object shapes, although limited visibility within the grip and shakiness in tasks like writing presented challenges. The Kwawu hand struggled with tasks requiring alignment and precision, which caused longer completion times and lower user preference. The Hosmer hook offered precision for fine motor tasks and was rated highly by participants for ease of use, but it had difficulty in grasping larger items.

\section{Future Work} 

Future iterations of the Everting hand should include transparent hydrostatic materials to address participant feedback on visual obstruction during tasks like stacking. Further work should also refine the hydrostat’s structure to reduce shakiness and improve control for tasks requiring fine motor skills, such as writing. It would be interesting to incorporate advanced sensors to monitor grip force in real time, as well as lightweight materials that could reduce the device’s weight for extended wearability. Testing with individuals with upper-limb differences would provide further insights into usability.

Our pilot user study primarily focused on handling smaller objects, and the user study results (Table~\ref{tab:Jebsen_fails}) suggest potential limitations with large and heavy items. To address this, future work should involve increasing the grasping area. Another important aspect of the Everting hand that we have not yet addressed is its durability. While the hydrostat's membrane was not damaged throughout the user study, sharp objects could potentially puncture it, leading to device failure. Therefore, future evaluations should assess the hydrostat's durability and investigate alternative materials with greater resilience for the membrane. Additionally, while the body-powered transmission system proved effective, it was not fully optimized. Ongoing research into refined transmission systems could present ways for improvement~\cite{abbott2023characterizing, mcpherson2023wearable}.

Another area of interest is user responses to the Everting hand’s non-anthropomorphic aesthetics. While prior studies suggest that users adapt to the functionality of unconventional designs in both physical and virtual environments~\cite{molnar2022toward}, further studies on user preferences for non-anthropomorphic aesthetics could provide valuable insights. Together, these enhancements could broaden the Everting hand’s applications and support its role as a practical prosthetic solution.

\section{Acknowledgements} \label{sec:Acknowledgement}
We thank the Notre Dame e-NABLE club for useful discussions about building the Kwawu hand, and the Notre Dame Engineering Innovation Hub for assistance with fabricating the Everting hand.

\bibliographystyle{IEEEtran}
\bibliography{library}

\begin{thebibliography}{10}
\providecommand{\url}[1]{#1}
\csname url@samestyle\endcsname
\providecommand{\newblock}{\relax}
\providecommand{\bibinfo}[2]{#2}
\providecommand{\BIBentrySTDinterwordspacing}{\spaceskip=0pt\relax}
\providecommand{\BIBentryALTinterwordstretchfactor}{4}
\providecommand{\BIBentryALTinterwordspacing}{\spaceskip=\fontdimen2\font plus
\BIBentryALTinterwordstretchfactor\fontdimen3\font minus \fontdimen4\font\relax}
\providecommand{\BIBforeignlanguage}[2]{{%
\expandafter\ifx\csname l@#1\endcsname\relax
\typeout{** WARNING: IEEEtran.bst: No hyphenation pattern has been}%
\typeout{** loaded for the language `#1'. Using the pattern for}%
\typeout{** the default language instead.}%
\else
\language=\csname l@#1\endcsname
\fi
#2}}
\providecommand{\BIBdecl}{\relax}
\BIBdecl

\bibitem{cordella2016literature}
F.~Cordella, A.~L. Ciancio, R.~Sacchetti, A.~Davalli, A.~G. Cutti, E.~Guglielmelli, and L.~Zollo, ``Literature review on needs of upper limb prosthesis users,'' \emph{Frontiers in Neuroscience}, vol.~10, no. 209, pp. 1--14, 2016.

\bibitem{stephens2019survey}
B.~Stephens-Fripp, M.~J. Walker, E.~Goddard, and G.~Alici, ``A survey on what {A}ustralians with upper limb difference want in a prosthesis: Justification for using soft robotics and additive manufacturing for customized prosthetic hands,'' \emph{Disability and Rehabilitation: Assistive Technology}, vol.~15, no.~3, pp. 342--349, 2019.

\bibitem{resnik2021longitudinal}
L.~Resnik, M.~Borgia, S.~Biester, and M.~A. Clark, ``Longitudinal study of prosthesis use in veterans with upper limb amputation,'' \emph{Prosthetics and Orthotics International}, vol.~45, no.~1, pp. 26--35, 2021.

\bibitem{kim2022influence}
J.~J. Kim, J.~Kim, J.~Lee, and J.~Shin, ``Influence of lifestyle pattern on preference for prosthetic hands: Understanding the development pathway for {3D}-printed prostheses,'' \emph{Journal of Cleaner Production}, vol. 379, p. 134599, 2022.

\bibitem{vertongen2020mechanical}
J.~Vertongen, D.~G. Kamper, G.~Smit, and H.~Vallery, ``Mechanical aspects of robot hands, active hand orthoses, and prostheses: A comparative review,'' \emph{IEEE/ASME Transactions on Mechatronics}, vol.~26, no.~2, pp. 955--965, 2020.

\bibitem{carrozza2004spring}
M.~C. Carrozza, C.~Suppo, F.~Sebastiani, B.~Massa, F.~Vecchi, R.~Lazzarini, M.~R. Cutkosky, and P.~Dario, ``The spring hand: development of a self-adaptive prosthesis for restoring natural grasping,'' \emph{Autonomous Robots}, vol.~16, no.~2, pp. 125--141, 2004.

\bibitem{dollar2006robust}
A.~M. Dollar and R.~D. Howe, ``A robust compliant grasper via shape deposition manufacturing,'' \emph{IEEE/ASME transactions on mechatronics}, vol.~11, no.~2, pp. 154--161, 2006.

\bibitem{kwawu_hand}
\BIBentryALTinterwordspacing
e\textminus NABLE~Community, ``Kwawu hand,'' Thingiverse, 2017, accessed: 2024-10-21. [Online]. Available: \url{https://www.thingiverse.com/thing:2774477}
\BIBentrySTDinterwordspacing

\bibitem{shintake2018soft}
J.~Shintake, V.~Cacucciolo, D.~Floreano, and H.~Shea, ``Soft robotic grippers,'' \emph{Advanced Materials}, vol.~30, no.~29, p. 1707035, 2018.

\bibitem{brown2010universal}
E.~Brown, N.~Rodenberg, J.~Amend, A.~Mozeika, E.~Steltz, M.~R. Zakin, H.~Lipson, and H.~M. Jaeger, ``Universal robotic gripper based on the jamming of granular material,'' \emph{Proceedings of the National Academy of Sciences}, vol. 107, no.~44, pp. 18\,809--18\,814, 2010.

\bibitem{nishida2016development}
T.~Nishida, Y.~Okatani, and K.~Tadakuma, ``Development of universal robot gripper using {MR} $\alpha$ fluid,'' \emph{International Journal of Humanoid Robotics}, vol.~13, no.~4, p. 1650017, 2016.

\bibitem{ilievski2011soft}
F.~Ilievski, A.~D. Mazzeo, R.~F. Shepherd, X.~Chen, and G.~M. Whitesides, ``Soft robotics for chemists,'' \emph{Angewandte Chemie}, vol. 123, no.~8, pp. 1930--1935, 2011.

\bibitem{hao2016universal}
Y.~Hao, Z.~Gong, Z.~Xie, S.~Guan, X.~Yang, Z.~Ren, T.~Wang, and L.~Wen, ``Universal soft pneumatic robotic gripper with variable effective length,'' in \emph{Chinese Control Conference}, 2016, pp. 6109--6114.

\bibitem{li2020bioinspired}
H.~Li, J.~Yao, C.~Liu, P.~Zhou, Y.~Xu, and Y.~Zhao, ``A bioinspired soft swallowing robot based on compliant guiding structure,'' \emph{Soft Robotics}, vol.~7, no.~4, pp. 491--499, 2020.

\bibitem{sui2022bioinspired}
D.~Sui, Y.~Zhu, S.~Zhao, T.~Wang, S.~K. Agrawal, H.~Zhang, and J.~Zhao, ``A bioinspired soft swallowing gripper for universal adaptable grasping,'' \emph{Soft Robotics}, vol.~9, no.~1, pp. 36--56, 2022.

\bibitem{root2021bio}
S.~E. Root, D.~J. Preston, G.~O. Feifke, H.~Wallace, R.~M. Alcoran, M.~P. Nemitz, J.~A. Tracz, and G.~M. Whitesides, ``Bio-inspired design of soft mechanisms using a toroidal hydrostat,'' \emph{Cell Reports Physical Science}, vol.~2, no.~9, p. 100572, 2021.

\bibitem{enable}
\BIBentryALTinterwordspacing
e\textminus NABLE~Community, ``{e\textminus NABLE}: {3D} printed prosthetics for the world,'' 2023, accessed: 2024-10-21. [Online]. Available: \url{https://enablingthefuture.org/}
\BIBentrySTDinterwordspacing

\bibitem{mathiowetz1985adult}
V.~Mathiowetz, G.~Volland, N.~Kashman, and K.~Weber, ``Adult norms for the box and block test of manual dexterity,'' \emph{The American Journal of Occupational Therapy}, vol.~39, no.~6, pp. 386--391, 1985.

\bibitem{jebsen1969objective}
R.~H. Jebsen, ``An objective and standardized test of hand function,'' \emph{Archives of Physical Medicine and Rehabilitation}, vol.~50, pp. 311--319, 1969.

\bibitem{openscad}
\BIBentryALTinterwordspacing
OpenSCAD, ``{OpenSCAD}: The programmable {3D} modeler,'' accessed: 2024-10-28. [Online]. Available: \url{https://www.openscad.org/}
\BIBentrySTDinterwordspacing

\bibitem{nasa_anthropometric_source_book}
NASA, \emph{Anthropometric Source Book, Volume I: Anthropometry for Designers}, 1978, {NASA} Technical Paper 1024.

\bibitem{abbott2023characterizing}
M.~E. Abbott and H.~S. Stuart, ``Characterizing the force-motion tradeoff in body-powered transmission design,'' \emph{IEEE Transactions on Neural Systems and Rehabilitation Engineering}, vol.~31, pp. 3064--3074, 2023.

\bibitem{mcpherson2023wearable}
A.~I. McPherson, M.~E. Abbott, W.~White, Y.~Gloumakov, and H.~S. Stuart, ``A wearable testbed for studying variable transmission in body-powered prosthetic gripping,'' in \emph{International Conference on Rehabilitation Robotics}, 2023, pp. 1--6.

\bibitem{molnar2022toward}
J.~Molnar and Y.~Menguc, ``Toward handling the complexities of non-anthropomorphic hands,'' in \emph{CHI Conference on Human Factors in Computing Systems Extended Abstracts}, 2022, pp. 1--9.

\end{thebibliography}

\end{document}